\documentclass[11pt]{article}

\usepackage[final]{acl}

\usepackage{times}
\usepackage{latexsym}

\usepackage[T1]{fontenc}

\usepackage[utf8]{inputenc}

\usepackage{microtype}

\usepackage{inconsolata}

\usepackage{graphicx}
\usepackage{booktabs}
\usepackage{multirow}        
\usepackage{colortbl}        
\usepackage{adjustbox}       
\usepackage{arydshln}    

\definecolor{mygray}{gray}{.9}

\usepackage{amssymb}         
\usepackage{pifont}          

\usepackage{amsmath,amssymb,amsfonts}
\usepackage[ruled,vlined, linesnumbered]{algorithm2e}
\usepackage[accsupp]{axessibility}  
\usepackage{cleveref}
\usepackage{subcaption}

\title{CLIMB: Confidence-Guided Complementary Evidence for Multimodal Retrieval-Augmented Generation}

\author{
Hang Gao$^{1}$ \quad
Wujiang Xu$^{1}$ \quad
Zhixing Zhang$^{2}$ \quad
Kai Mei$^{1}$ \\
\textbf{Jingyi Yang}$^{2}$ \quad
\textbf{Dimitris N. Metaxas}$^{1}$ \\[0.1em]
$^{1}$Rutgers University \quad $^{2}$Google\\
\small{
    \textbf{Correspondence:}
    \href{mailto:h.gao@rutgers.edu}{h.gao@rutgers.edu},
    \href{mailto:dnm@cs.rutgers.edu}{dnm@cs.rutgers.edu}
}
}

\begin{document}
\maketitle
\begin{abstract}
Multimodal large language models (MLLMs) have shown strong visual reasoning abilities, but knowledge-intensive visual question answering often requires external textual evidence beyond the image and the model's parametric knowledge. Existing multimodal RAG systems commonly rely on Top-$K$ retrieval or reranking, which may return redundant passages and provide limited control over whether an answer update is sufficiently supported by the retrieved evidence. We propose \textit{CLIMB}, a training-free inference-time framework for multimodal RAG. CLIMB first constructs a compact complementary evidence pool using an MMR-style objective that balances query relevance and passage-level redundancy. It then performs confidence-controlled refinement within this fixed pool: an R/E/C critic scores passages by relevance, evidence specificity, and cross-modal alignment, while an evidence-grounded confidence estimator accepts an updated answer only when the estimated confidence increases. This design provides a simple stopping criterion and reduces unnecessary refinement without modifying the underlying retriever or MLLM. Experiments on Encyclopedic-VQA and InfoSeek show that CLIMB consistently improves over retrieval-augmented multimodal baselines. Ablations further indicate that complementary pooling, critic-based scoring, and iterative confidence-controlled refinement each contribute to the final performance.
\end{abstract}

\section{Introduction}
\begin{figure}[t]
  \centering
    \includegraphics[width=\columnwidth]{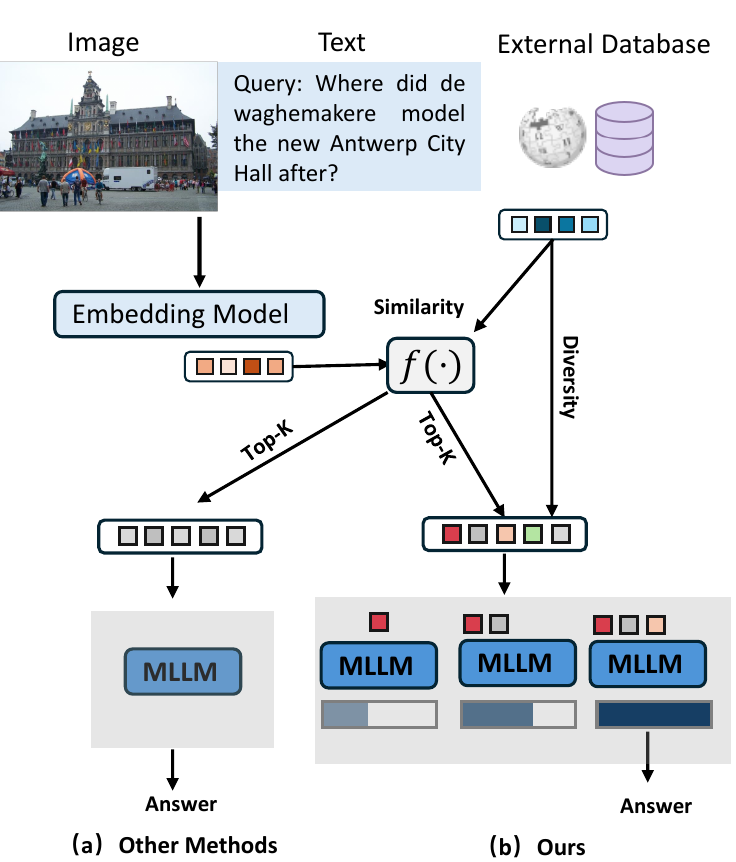}
    \caption{Comparison of multimodal RAG approaches. (a) Standard methods \cite{cocchi2025augmenting,caffagni2024wiki,yan2024echosight} rely on Top-K similarity selection, which often produces redundant evidence with limited coverage and lacks explicit quality control for answer generation. (b) Our CLIMB framework addresses these limitations through two key innovations: diversity-aware evidence construction that balances relevance and complementarity to build a non-redundant passage pool, and confidence-controlled decoding that uses multi-faceted scoring and monotonic confidence constraints to ensure reliable answer updates.}
    \vspace{-22pt}
    \label{toyexample}
\end{figure}
Multimodal large language models (MLLMs) \cite{llava1.5,li2024llava,abdin2024phi} have made rapid progress in visual reasoning and grounded dialogue, yet knowledge-intensive visual question answering remains challenging \cite{caffagni2024revolution, wu2023multimodal, yin2024survey}. In these tasks, the image provides important grounding cues, but the answer often depends on external facts such as entity attributes, historical details, geographic information, or fine-grained taxonomy. Such information is difficult to cover fully through parametric knowledge alone, especially for long-tail entities and time-sensitive facts \cite{ye2024mplug, llava1.5, li2023blip, chen2023can, mensink2023encyclopedic}.

Retrieval-augmented generation (RAG) addresses this limitation by providing external textual evidence at inference time \cite{lewis2020retrieval, zhao2023retrieving, gao2024vector, liu2025hm}. In multimodal RAG, however, performance depends not only on retrieving relevant passages, but also on assembling a compact evidence set that is complementary, non-redundant, and sufficiently reliable for answer generation. A retrieval set that contains several near-duplicate passages may have high similarity scores but still fail to cover the specific facets needed to answer the question.

Despite steady advances, common multimodal RAG pipelines exhibit limitations rooted in design choices. The prevailing practice ranks passages by image-text similarity and conditions the MLLM on the top-ranked set. Such Top-K assembly concentrates scores on near-duplicates \cite{yan2024echosight}, producing redundant contexts with limited semantic coverage, reflecting the core challenge of \textit{evidence complementarity}. Enhancements in retrievers and hierarchical search reduce the candidate space \cite{caffagni2024wiki}, yet the final assembly step rarely suppresses pairwise redundancy in a principled way. 
Iterative RAG variants \cite{arefeen2024irag} attempt to address insufficient evidence through multiple retrieval-and-rewrite cycles, but face the challenges of \textit{quality control} and \textit{termination criteria}: these systems increase latency and variance while offering limited transparency about when updates should be trusted or when to stop refinement. Most architectures also lack explicit mechanisms for reliability assessment, leaving systems unable to determine when evidence is sufficient. These issues motivate an inference-time mechanism that jointly considers evidence complementarity and answer-level verifiability.

To tackle these challenges, we propose \textit{CLIMB}, a framework that treats multimodal RAG as two coupled decisions: which evidence should be assembled, and when an answer update should be accepted. CLIMB first builds a fixed complementary micro-pool using an MMR-style objective that balances query relevance with redundancy suppression. This pool provides a compact set of candidate passages that are individually relevant while covering different aspects of the question. Within this fixed pool, CLIMB performs confidence-controlled refinement. A passage-level R/E/C critic scores each candidate by relevance, evidence specificity, and cross-modal alignment. The model then generates an answer from the selected subset, estimates an evidence-grounded confidence score, and accepts a revised answer only if the estimated confidence improves. Because all refinement steps operate inside the fixed pool, CLIMB avoids repeated external retrieval and provides a bounded, interpretable stopping rule. Empirically, CLIMB delivers consistent gains on knowledge-intensive VQA benchmarks. Across datasets such as Encyclopedic-VQA\cite{mensink2023encyclopedic} and InfoSeek\cite{chen2023can}, the framework improves over Top-$K$ retrieval and iterative RAG variants, and it preserves performance on instances where additional evidence is unnecessary. Ablations indicate that critic is a primary driver of coverage and answer quality, while confidence-controlled acceptance contributes to stability, calibration, and early stopping. Our contributions are summarized as follows:
\begin{enumerate}
    \item We present CLIMB, which constructs a complementary evidence pool via an MMR-style objective that promotes relevance while suppressing redundancy.
    \item Building on this pool, we introduce confidence-controlled decoding with an R/E/C scoring interface and an acceptance rule that updates only when a scalar confidence increases, enabling monotone improvement and natural early stopping.
    \item We systematize multimodal RAG as a complementarity-aware problem, providing a modular pipeline that integrates with standard encoders and retrievers and aligns with hierarchical retrieval and reflective-control designs.
\end{enumerate}

\section{Related Work}
\label{sec:relatedwork}
Multimodal large language models (MLLMs) extend text-only LLMs by coupling a vision encoder with a language backbone, enabling grounded question answering, visual dialogue, and open-ended multimodal generation \cite{radford2021learning, liu2023visual, mckinzie2024mm1, sun2024emu}. Architecturally, systems differ in how they bridge modalities, from query-former style adaptors and projection layers to end-to-end finetuning, and in how they leverage instruction tuning and conversational data for alignment \cite{radford2021learning, llava1.5, lerner2024cross, qi2024rora}. Subsequent work improves robustness and utility through better pretraining corpora, mixed-format curricula, and task-oriented instruction sets, yielding models that can act as general-purpose perceptual reasoners under open-world conditions \cite{alayrac2022flamingo, changpinyo2021conceptual, gadre2023datacomp, deitke2024molmo}.

Another line of research showed great potential of Retrieval-augmented generation (RAG), which injects non-parametric evidence into the context to expand coverage and mitigate hallucination, first established in text settings by combining sparse or dense retrievers with sequence generators \cite{lewis2020retrieval, guu2020retrieval, izacard2023atlas, borgeaud2022improving, gao2026beyond}. Early systems such as Fusion-in-Decoder and subsequent large-scale variants demonstrated that learned retrieval over web-scale corpora, together with retrieval-conditioned decoding, substantially improves knowledge-intensive QA \cite{gao2023retrieval, arslan2024survey, fan2024survey, gao2026pooling}. As instruction-tuned LLMs became the default interface, RAG shifted toward inference-time augmentation: passages are formatted into input sequences, sometimes routed or filtered by reflection and verifier modules that decide whether to retrieve, how much to use, and when to revise, with attention to calibration and reproducibility of the retrieval–decoding loop \cite{asai2024self, yang2025knowing}. Extending RAG to multimodal LLMs introduces image-conditioned retrieval and multimodal context assembly. Recent pipelines structure candidates hierarchically, coarse selection guided by image or title cues followed by question- and text-level refinement, while keeping the MLLM backbone intact \cite{caffagni2024wiki, yan2024echosight}. Other approaches integrate reflective signals into decoding to regulate access to external knowledge for knowledge-based VQA \cite{cocchi2025augmenting}. These developments collectively move from simply “finding relevant text” toward assembling an image- and question-aware context that a multimodal decoder can reliably condition on. However, assembling a compact, complementary context and governing acceptance by answer reliability remain under-specified at inference time. Our CLIMB framework targets this gap by pairing an explicitly complementary, fixed evidence pool with confidence-governed decoding, providing a protocol for both diversity-aware assembly and monotone, reliability-driven refinement on top of standard multimodal RAG pipelines.

\section{Method}
\label{sec:method}
\begin{figure*}[t]
  \centering
    \includegraphics[width=\linewidth]{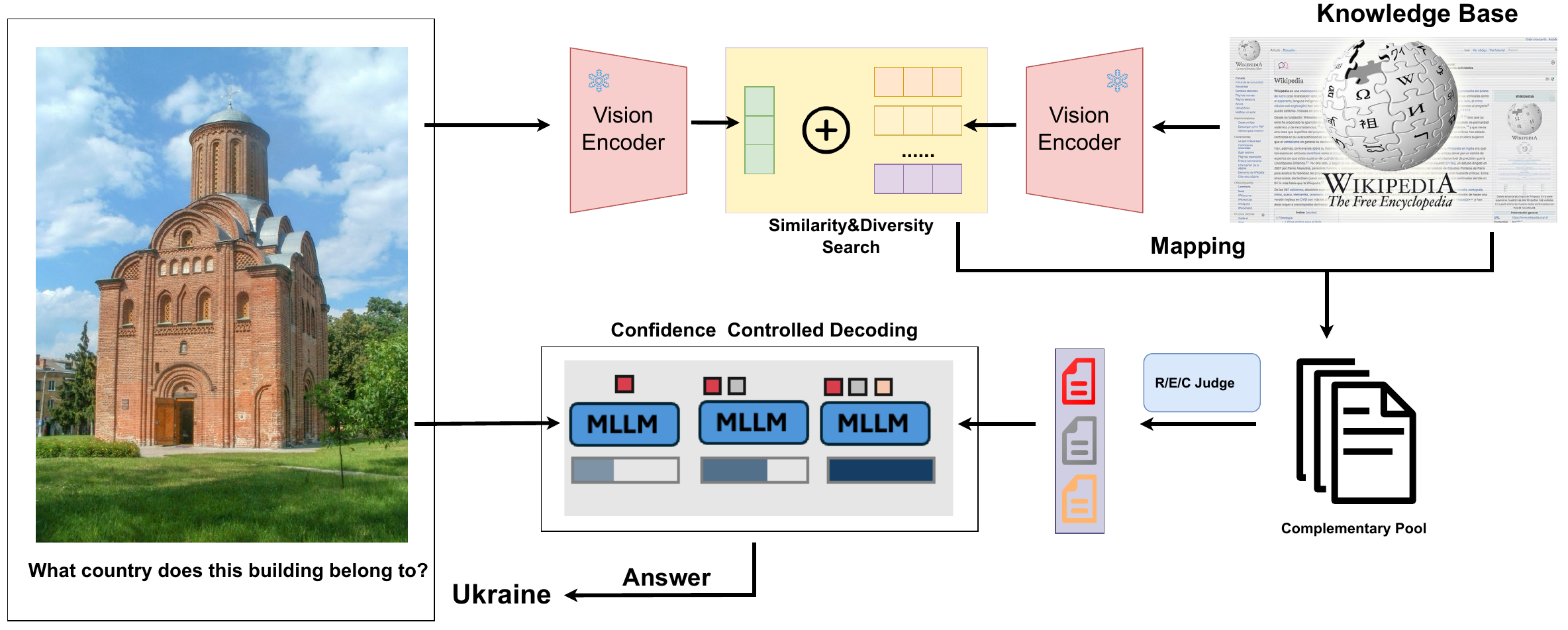}
    \caption{\textbf{Overview of \textit{CLIMB}.} Given an image and a question, CLIMB retrieve a small candidate set from the knowledge base and build a complementary evidence pool via similarity–diversity selection. An R/E/C critic scores passages, and a frozen MLLM generates answers from the top-$M$ evidence while a confidence estimator and gap diagnosis drive an iterative refinement loop, accepting updates only when confidence increases.}
    \label{workflow}
\end{figure*}
CLIMB is an inference-time framework for multimodal RAG with two stages: complementarity-aware evidence assembly and confidence-controlled decoding. Given an image, a question, and an external corpus, CLIMB first builds a compact fixed-size evidence pool. It then iteratively selects evidence subsets and accepts an updated answer only when the estimated evidence-grounded confidence improves. The base MLLM is kept frozen and treated as a black-box generator. Figure \ref{workflow} provides an overview.

\subsection{Problem Formulation}
\label{subsec:formulation}
We consider knowledge-intensive visual question answering. Each example consists of an image $I$, a question $q$, and an external text corpus $\mathcal{D}$. The goal is to generate an answer $a$ supported by a subset of passages $\mathcal{S} \subset \mathcal{D}$, and a scalar confidence $U\in[0,1]$ indicating how well the answer can be verified. CLIMB treats the MLLM as a frozen generator and focuses on inference-time evidence selection. Instead of repeatedly querying the full corpus during refinement, CLIMB first constructs a fixed complementary micro-pool $\mathcal{P} \subset \mathcal{D}$ of size $K$. All subsequent scoring, facet diagnosis, and answer refinement operate within $\mathcal{P}$.

\subsection{Complementary Evidence Pool}
\label{subsec:complementarity}

Vision--text retrieval typically ranks passages by similarity to the image or the query, often surfacing a cluster of near-duplicates at the top. Such redundancy limits semantic coverage and leaves later reasoning steps without the diverse evidence needed for knowledge-intensive VQA. \textit{CLIMB} therefore constructs a compact \emph{micro-pool} of size $K$ whose members are individually relevant yet mutually non-overlapping.

Given an image $I$ and question $q$, a standard CLIP-based search retrieves a candidate set $\mathcal{C}_N \subset \mathcal{D}$. We reuse the retriever’s similarity as a query–passage relevance score $\mathrm{rel}(I,q,t)$ for each $t\in\mathcal{C}_N$. From the screened candidates, we assemble a micro-pool $\mathcal{P}$ using a Maximal Marginal Relevance (MMR) \cite{carbonell1998use} objective. Let $S$ denote the set of already admitted items, and let $\mathrm{sim}(t,u)$ be cosine similarity between the text embeddings of $t$ and $u$. Starting from the most relevant passage, each subsequent passage is chosen by
\begin{equation}
\label{eq:mmr}
\begin{aligned}
t^{\star}
=
\arg\max_{t\in \mathcal{C}_N\setminus S}
\{&
\lambda\,\mathrm{rel}(I,q,t) \\
&-
(1-\lambda)\max_{u\in S}\mathrm{sim}(t,u)
\}.
\end{aligned}
\end{equation}
where $\lambda \in [0,1]$ balances query fit and redundancy (default $\lambda=0.6$). Penalizing the maximum similarity discourages the strongest duplication, so that each admitted passage is encouraged to contribute a distinct facet of the question.

With cached embeddings, this one-pass selection costs $O(NK)$ for a candidate set of size $N$ and a pool of size $K$. The resulting micro-pool is fixed for the remaining refinement steps. This design keeps the evidence search space bounded and reduces variation caused by repeated external retrieval. We set $K=5$ by default and analyze its effect in the sensitivity study.

\subsection{Evidence Judge and Confidence Control}
\label{subsec:evidence_confidence}
Once the micro-pool $\mathcal{P}$ is fixed, CLIMB evaluates evidence at the passage level and answer level.
\paragraph{R/E/C evidence judge.}
For each passage $t \in \mathcal{P}$, CLIMB assigns three interpretable scores. The relevance score $R \in \{1,\ldots,5\}$ measures how directly $t$ addresses the question. The evidence specificity score $E \in \{1,\ldots,5\}$ measures whether $t$ contains concrete and verifiable details, such as named entities, dates, numbers, or definitions. The cross-modal alignment score $C \in \{1,\ldots,5\}$ measures whether $t$ is consistent with the visual content of $I$. These scores are produced by a constrained judgment prompt to the frozen MLLM. After normalization to $[0,1]$, they are combined as
\begin{equation}
\phi(t) = \alpha_R \tilde{R}(t) + \alpha_E \tilde{E}(t) + \alpha_C \tilde{C}(t),
\end{equation}
where $\alpha_R$, $\alpha_E$, and $\alpha_C$ are non-negative weights that sum to one. We use equal weights by default. The initial evidence subset is selected as
\begin{equation}
\mathcal{S}_0 = \operatorname*{arg\,topM}_{t \in \mathcal{P}} \phi(t), 
\qquad M = \lceil \sqrt{K} \rceil .
\end{equation}

\paragraph{Verifiability-oriented confidence.}
Given an answer $a$ generated from an evidence subset $\mathcal{S}$, CLIMB estimates a scalar confidence score $U \in [0,1]$ based on three bounded signals: coverage, consistency, and specificity. Coverage measures whether the main factual claims in $a$ are supported by passages in $\mathcal{S}$. Consistency penalizes conflicts among the selected passages or between the passages and the image. Specificity rewards answers that contain concrete and verifiable information. The three scores are averaged to obtain $U$. In the iterative refinement procedure that follows, this confidence is used relatively: a new answer is accepted only if its confidence strictly increases, and refinement stops early once $U$ exceeds a target threshold $U_{\mathrm{STAR}}$ or a maximum number of rounds is reached. This confidence score is not interpreted as a calibrated probability of correctness, nor do we treat the critic as an independent external oracle. Instead, confidence is used as a relative, evidence-grounded control signal within the fixed micro-pool. The estimator is constrained to evaluate whether the factual claims in the answer are covered by the selected passages, whether the answer is consistent with the image and retrieved evidence, and whether the answer is sufficiently specific to be verified. 

\subsection{Facet Diagnosis and Pool-Internal Refresh}
\label{subsec:facets}

Some errors arise when the current subset $\mathcal{S}_t$ misses a key facet of the question. Rather than performing another external retrieval, \textit{CLIMB} diagnoses under-covered facets and refreshes evidence \emph{within} the fixed micro-pool $\mathcal{P}$.

Given the current state $(I,q,\mathcal{S}_t,a_t)$, we elicit at most two short facet phrases
\[
\mathcal{G}_t = \{g_1,g_2\},
\]
each describing a missing aspect that prevents full support (we constrain each $g_i$ to be a short phrase). If no meaningful facet can be identified, refinement terminates. For every passage $u\in\mathcal{P}$, we then compute a lightweight facet–text match score $\psi(u \mid \mathcal{G}_t)$ using a combination of lexical overlap (e.g., BM25) and sentence-level embedding similarity. The original evidence scores $\phi(u)$ are updated as
\begin{equation}
\label{eq:phi_refresh}
\phi'(u)
=
\phi(u)
+
\gamma\,\psi(u \mid \mathcal{G}_t),
\qquad
\gamma \in [0,1],
\end{equation}
and the next working subset $\mathcal{S}_{t+1}$ is chosen as the top-$M$ passages under $\phi'$.

\textit{CLIMB }then queries the frozen MLLM once with $(I,q,\mathcal{S}_{t+1})$ to obtain a new answer $a_{t+1}$ and compute its confidence $U_{t+1}$. The update is accepted only if $U_{t+1} > U_t$; otherwise the procedure stops and returns $(a_t,U_t,\mathcal{S}_t)$. In practice we also cap the number of refinement rounds by a small constant $T_{\max}$ (we use $T_{\max}=3$) and stop early once $U_t \ge U_{\mathrm{STAR}}$. Because all operations remain inside the fixed micro-pool, \textit{CLIMB} avoids re-retrieval, keeps latency predictable, and enforces a monotone improvement in verifiability across accepted updates.

\subsection{Algorithm}
\label{subsec:algorithm}
\begin{algorithm}[t]
\caption{CLIMB}
\label{alg:climb}

\KwIn{Image $I$, Question $q$, Corpus $\mathcal{D}$}
\KwOut{Answer $a$, Confidence $U$, Supporting $\mathcal{S}$}

\BlankLine
\textbf{Candidate Retrieval}\;
$\mathcal{C}_N \leftarrow \textsc{Retrieve}(I,q,\mathcal{D})$\;
$\mathcal{C}_N \leftarrow \textsc{FilterByRel}(\mathcal{C}_N,\tau)$\;

\BlankLine
\textbf{Complementary Pool}\;
$\mathcal{P} \leftarrow \textsc{BuildMicroPool}(I,q,\mathcal{C}_N,K)$\;  

\BlankLine
\textbf{Initial Scoring and Answer}\;
$\phi \leftarrow \textsc{ScoreREC}(I,q,\mathcal{P})$\;
$\mathcal{S}_0 \leftarrow \textsc{TopM}(\mathcal{P},\phi,M)$\;
$(a_0,U_0) \leftarrow \textsc{GenerateAndScore}(I,q,\mathcal{S}_0)$\;

\BlankLine
\textbf{Iterative Confidence Climb}\;
\For{$t = 0$ \KwTo $T_{\max}-1$}{
    \If{$U_t \ge U_{\mathrm{STAR}}$}{
        \textbf{break}\;
    }
    $\mathcal{G}_t \leftarrow \textsc{DiagnoseFacets}(I,q,\mathcal{S}_t,a_t)$\;
    \If{$\mathcal{G}_t = \emptyset$}{
        \textbf{break}\;
    }

    $\phi' \leftarrow \textsc{RefreshScores}(\mathcal{P},\phi,\mathcal{G}_t,\gamma)$\;
    $\mathcal{S}_{t+1} \leftarrow \textsc{TopM}(\mathcal{P},\phi',M)$\;
    $(a_{t+1},U_{t+1}) \leftarrow \textsc{GenerateAndScore}(I,q,\mathcal{S}_{t+1})$\;

    \If{$U_{t+1} \le U_t$}{
        \textbf{break}\;
    }
}
\BlankLine
\Return{$a_t,\, U_t,\, \mathcal{S}_t$}\;
\end{algorithm}

The end-to-end procedure of \textit{CLIMB} is summarized in Algorithm \ref{alg:climb}. The algorithm take $(I,q,\mathcal{D})$ as the input and output the answer, confidence and supporting set. Lines 2-3 perform a small vision–text retrieval followed by lightweight relevance filtering to obtain a candidate set $\mathcal{C}_N$. Lines 4-5 construct a complementary micro-pool $\mathcal{P}$ of size $K$ via \textsc{BuildMicroPool}, which implements the MMR objective in Eq.~\ref{eq:mmr} to balance query relevance against pairwise redundancy. In lines 7-9, passages in $\mathcal{P}$ are scored by the R/E/C critic to produce $\phi$, the top-$M$ passages form the initial subset $\mathcal{S}_0$, and a single call to the frozen MLLM yields the initial answer and confidence $(a_0,U_0)$. Lines 11-16 implement the confidence-guided refinement. At each iteration, \textit{CLIMB} first checks whether the current confidence has already exceeded the target threshold $U_{\mathrm{STAR}}$; if so, the procedure terminates. It then diagnoses up to two under-supported facets $\mathcal{G}_t$ from the current state $(I,q,\mathcal{S}_t,a_t)$; if no facet is found, refinement stops. Otherwise, lines 17-19 refresh the evidence scores inside the fixed pool using \textsc{RefreshScores}, select a new subset $\mathcal{S}_{t+1}$, and obtain an updated answer and confidence $(a_{t+1},U_{t+1})$. The update is accepted only when $U_{t+1}>U_t$; if confidence fails to increase, the loop exits and returns the last accepted $(a_t,U_t,\mathcal{S}_t)$. This fixed-pool, monotone confidence climb avoids retrieval drift, stabilizes behavior across runs.

\section{Experiments}
\begin{table*}[t]
\begin{adjustbox}{width=1\textwidth,center}
\begin{tabular}{ll||c||cc||ccc} 
\toprule
\rowcolor{mygray}
\textbf{Model} & \textbf{LLM} & \textbf{Retrieval Mode} & \multicolumn{2}{c||}{\textbf{E-VQA}} & \multicolumn{3}{c}{\textbf{InfoSeek}} \\
\rowcolor{mygray}
& & & \textbf{Single-Hop} & \textbf{All} & \textbf{Unseen-Q} & \textbf{Unseen-E} & \textbf{All} \\
\hline \hline
\multicolumn{2}{l||}{\textit{Zero-shot LLMs}} & & & & & & \\
Vanilla & Vicuna-7B & - & 2.1 & 2.0 & 0.3 & 0.0 & 0.0 \\
Vanilla & LLaMA-3.1-8B & - & 16.5 & 16.6 & 2.1 & 0.0 & 0.0 \\
\hline
\multicolumn{2}{l||}{\textit{Zero-shot Multimodal LLMs}} & & & & & & \\
LLaVA-v1.5 & Vicuna-7B & - & 16.3 & 16.9 & 9.6 & 9.4 & 9.5 \\
LLaVA-v1.5 & LLaMA-3.1-8B & - & 16.0 & 16.9 & 8.3 & 8.9 & 7.8 \\
\hline
\multicolumn{2}{l||}{\textit{Retrieval-Augmented Models}} & & & & & & \\
DPR$_\text{V+T}$ \textcolor{lightgray}{\scriptsize{[ECIR24\cite{lerner2024cross}]}} & Multi-passage BERT & CLIP ViT-B/32 & 29.1 & - & - & - & 12.4 \\
RORA-VLM \textcolor{lightgray}{\scriptsize{[arXiv24\cite{qi2024rora}]}} & Vicuna-7B & CLIP+Google Search & - & 20.3 & 25.1 & 27.3 & - \\
Wiki-LLaVA \textcolor{lightgray}{\scriptsize{[CVPRW24\cite{caffagni2024wiki}]}} & LLaMA-3.1-8B & CLIP ViT-L/14 & 18.3 & 19.6 & 28.6 & 25.7 & 27.1 \\
EchoSight \textcolor{lightgray}{\scriptsize{[EMNLP24\cite{yan2024echosight}]}} & LLaMA-3.1-8B & EVA-CLIP-8B & 22.4 & 21.7 & 30.0 & 30.7 & 30.4 \\
ReflectiVA \textcolor{lightgray}{\scriptsize{[CVPR25\cite{cocchi2025augmenting}]}} & LLaMA-3.1-8B & CLIP ViT-L/14 & 24.9 & 26.7 & 34.5 & 32.9 & 33.7 \\
\rowcolor{mygray}
$\blacktriangleright$ \textbf{CLIMB} & LLaMA-3.1-8B & CLIP ViT-L/14 & \textbf{31.3} & \textbf{34.8} & \textbf{39.7} & \textbf{38.5} & \textbf{39.1} \\
\hline
\end{tabular}
\end{adjustbox}
\caption{Accuracy on the Encyclopedic-VQA test set and the InfoSeek validation set.}
\label{table:mainresults}
\end{table*}
\subsection{Datasets}

We evaluate CLIMB on two knowledge-intensive VQA benchmarks that require combining image understanding with external Wikipedia knowledge.

\paragraph{Encyclopedic-VQA.}
Encyclopedic-VQA contains image--question pairs grounded in Wikipedia entities. Questions target fine-grained factual properties such as species attributes, geographic information, and historical facts. Following prior work, we use the standard single-hop setting and report the BERT matching score (BEM).

\paragraph{InfoSeek.}
InfoSeek contains fact-seeking visual questions over Wikipedia entities, including both automatically generated and human-authored examples. Compared with Encyclopedic-VQA, it covers broader entity types and question templates. We follow the official split and evaluation protocol, restrict retrieval to the provided knowledge base, and report VQA accuracy and relaxed accuracy when applicable.

\subsection{Baselines}
We compare CLIMB with three groups of baselines. First, we include text-only LLMs, Vicuna-7B \cite{vicuna} and LLaMA-3.1-8B \cite{llama3}, which receive only the question and have no access to the image or external corpus. Second, we evaluate zero-shot multimodal LLMs, including LLaVA-v1.5 \cite{llava1.5} with Vicuna-7B and LLaMA-3.1-8B backbones, which use the image but do not retrieve external knowledge. Third, we compare with retrieval-augmented multimodal systems, including DPRV+T \cite{lerner2024cross}, RORA-VLM \cite{qi2024rora}, Wiki-LLaVA \cite{caffagni2024wiki}, EchoSight \cite{yan2024echosight}, and ReflectiVA \cite{cocchi2025augmenting}. These baselines use external knowledge through retrieval or filtering and therefore provide the most relevant comparison for evaluating CLIMB's complementary evidence assembly and confidence-controlled refinement.

\subsection{Implementation Details}
We use CLIP ViT-L/14@336px as the visual encoder to obtain image embeddings and image--text similarity scores for retrieval, and LLaMA-3.1-8B model as the answer generator. For each query, we first retrieve $N{=}64$ candidate passages from the dataset-specific Wikipedia collection using cosine similarity in the CLIP embedding space, and then construct a complementary evidence pool of size $K{=}5$ with the similarity\&diversity objective. On top of this pool, we compute relevance, evidence specificity, and cross-modal alignment scores for all passages and calibrate them on a held-out development split to obtain a unified ranking signal. During decoding, the generator is conditioned on the top passages via a fixed structured instruction template, and our confidence module evaluates each candidate answer using coverage completeness, inter-evidence and image consistency, and answer specificity. Facet diagnosis and pool-internal refresh are applied for at most $T_{\max}{=}3$ rounds, and an updated answer is accepted only when it yields a strictly higher calibrated confidence score. All results are obtained under the knowledge-base-restricted protocol of each benchmark.

\subsection{Overall Effectiveness}
\textbf{Main results.} 
Table \ref{table:mainresults} reports accuracy on Encyclopedic-VQA and InfoSeek. 
Text-only LLMs provide a very weak baseline, and zero-shot multimodal LLMs improve mainly by grounding the question in the image, but still stay around 17\% on Encyclopedic-VQA and below 10\% on InfoSeek. This shows that visual grounding and parametric knowledge alone are insufficient for encyclopedic VQA over large knowledge bases. Retrieval-augmented systems narrow this gap: Wiki-LLaVA, EchoSight, and ReflectiVA all benefit from access to Wikipedia, lifting accuracy into the 20--30\% range and confirming the importance of explicit evidence retrieval. EchoSight uses a stronger retriever (EVA-CLIP-8B) than Wiki-LLaVA but does not consistently outperform ReflectiVA, suggesting that retrieval strength alone is not enough when the assembled contexts remain redundant. \textit{CLIMB} achieves the best performance across all evaluation splits and consistently improves over the strongest retrieval baseline. On Encyclopedic-VQA, it yields absolute gains of 6.4 on the Single-Hop subset and 8.1 on the All setting. The gains on Unseen-Q and Unseen-E are 5.2 and 5.6, respectively, indicating that \textit{CLIMB} not only retrieves useful passages but also regulates how they are selected. On InfoSeek, \textit{CLIMB} further improves accuracy by 5.4 over the best baseline, despite the different knowledge base and question style. These results suggest that complementary evidence construction and confidence-controlled decoding provide benefits beyond those obtained by stronger retrievers alone.

\subsection{Ablation Study}
\label{subsec:ablation}
\begin{table}[h]
\centering
\resizebox{0.85\linewidth}{!}{
\begin{tabular}{lcc}
\toprule
\textbf{Method} & \textbf{E-VQA} & \textbf{InfoSeek} \\
\midrule
\textbf{Full} & \textbf{34.8} & \textbf{39.1} \\
w/o Facet Diagnosis & 33.7 & 38.7 \\
w/o Complementary Evidence Pool & 32.4 & 38.1 \\
w/o Critic & 26.9 & 33.4 \\
w/o Iteration & 25.8 & 33.0 \\
\bottomrule
\end{tabular}
}
\caption{
Ablation study results on InfoSeek and E-VQA. 
}
\label{tab:ablationn}
\end{table}

Table \ref{tab:ablationn} reports the contribution of each component in \textit{CLIMB}.Removing the facet diagnosis step produces a small but consistent drop on both datasets, indicating that explicitly identifying under-supported facets helps the model redirect attention within the fixed pool. Eliminating the complementary pool (MMR) leads to a larger degradation, showing that suppressing redundancy is important for assembling evidence that covers multiple attributes of the queried entity. The critic, responsible for R/E/C scoring and confidence estimation, is the most influential component: without it, accuracy decreases sharply on both benchmarks, confirming that a notion of verifiability is important for filtering misleading passages and stabilizing decoding. Finally, disabling iterative refinement yields the lowest performance among all ablations, demonstrating that many queries require progressively incorporating additional facets that are not captured in a single pass. Together, these results highlight that complementary evidence construction and confidence-guided refinement jointly enable the robustness of \textit{CLIMB}.

\subsection{Confidence Binning Analysis}
\begin{table}[h]
\centering
\begin{tabular}{lc}
\hline
\textbf{Confidence bin} & \textbf{Accuracy} \\
\hline
$U < 0.70$ & 32.1 \\
$0.70 \leq U < 0.80$ & 34.3 \\
$0.80 \leq U < 0.90$ & 38.7 \\
$U \geq 0.90$ & 40.2 \\
\hline
\end{tabular}
\caption{Confidence-binned accuracy on InfoSeek under the default CLIMB setting.}
\label{tab:confidence-binning}
\end{table}
Since CLIMB relies on an MLLM-based confidence estimator to control answer refinement, we further analyze whether the estimated confidence is informative about answer correctness. We stress that CLIMB does not use confidence as an absolute probability of correctness. Instead, the confidence score is used only as a relative, evidence-grounded acceptance signal: an updated answer is accepted only when its estimated verifiability under the selected evidence improves.

Table~\ref{tab:confidence-binning} reports a confidence-binned analysis on InfoSeek under the default CLIMB setting. Higher-confidence predictions are consistently associated with higher empirical accuracy, increasing from 32.1\% for $U<0.70$ to 40.2\% for $U \geq 0.90$. This trend does not imply perfect calibration, but it suggests that the confidence estimator provides a useful reliability ranking signal for controlling refinement. The result is also consistent with the threshold sensitivity in Figure~\ref{fig:disadvantage}(a), where overly low thresholds accept less reliable updates, while overly strict thresholds can become conservative.

\subsection{Sensitivity Analysis}
\begin{figure*}[t]
    \centering
    \includegraphics[width=1\linewidth]{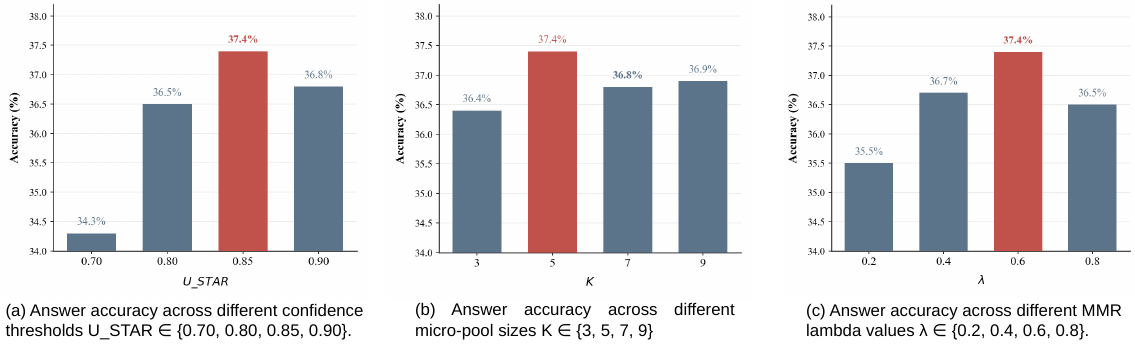}
    \caption{Sensitivity scores for hyperparameters $U_{STAR}$, $K$, $\lambda$, measured as the accuracy range across different parameter settings.}
    \label{fig:disadvantage}
\end{figure*}
\begin{figure}[t]
    \centering
    \begin{subfigure}[t]{0.45\columnwidth}
        \centering
        \includegraphics[width=\linewidth]{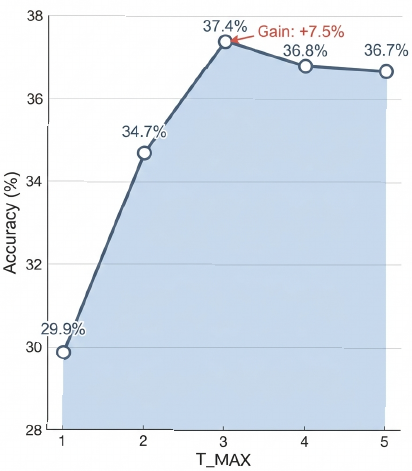}
        \caption{}
        \label{fig:iter}
    \end{subfigure}
    \hfill
    \begin{subfigure}[t]{0.53\columnwidth}
        \centering
        \includegraphics[width=\linewidth]{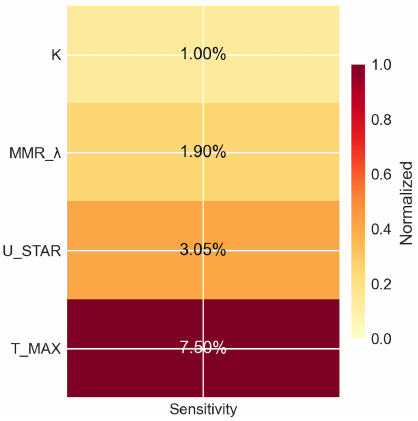}
        \caption{}
        \label{fig:sensitivity}
    \end{subfigure}
    \caption{Hyperparameter sensitivity of \textit{CLIMB}.
(a) Accuracy across different maximum iteration $T_{\max}$.
(b) Aggregated sensitivity scores across key hyperparameters, indicating that the framework is stable within a broad range of settings.}
 \vspace{-18pt}
    \label{fig:side_by_side}
\end{figure}
We further investigate how \textit{CLIMB} behaves under variations of its main hyperparameters by using the dataset InfoSeek: the micro-pool size $K$, the MMR trade-off $\lambda$, the confidence threshold $U_{\mathrm{STAR}}$, and the maximum number of refinement rounds $T_{\max}$. Figure \ref{fig:disadvantage} and  Figure \ref{fig:side_by_side}(a) summarizes results on InfoSeek. Varying the pool size $K$ reveals the expected coverage–noise trade-off. Very small pools lead to noticeable degradation, as the evidence set cannot cover all facets of the query. As $K$ increases, accuracy improves and then remains stable once a small complementary set is formed. Pushing $K$ further brings little or no gain and can slightly hurt performance, reflecting the fact that additional passages are often redundant or weakly relevant. The trade-off parameter $\lambda$ shows a similar pattern: accuracy is highest in a broad middle range where relevance and redundancy suppression are balanced, and deteriorates when $\lambda$ approaches extremes that either ignore diversity or over-penalize similar but useful passages. These trends are consistent across dataset, and the default configuration $(K{=}5,\lambda{=}0.6)$ lies well inside the stable region. The confidence threshold $U_{\mathrm{STAR}}$ primarily controls when the refinement loop accepts an updated answer. Lower thresholds allow less updates, while higher thresholds enforce more conservative behavior; in practice, accuracy remains stable across a wide interval around the default choice, indicating that the verifiability signal is reasonably calibrated and does not require precise tuning. The maximum number of refinement rounds $T_{\max}$ has a limited effect beyond moving from a single pass to a small number of iterations: performance improves when allowing more than one round, then saturates, and additional iterations bring negligible changes, which matches the observation that most queries require at most two or three refinement steps.

To summarize the overall stability, we also report a sensitivity score that measures the accuracy range induced by sweeping each hyperparameter within the tested bounds in Figure \ref{fig:side_by_side}(b). The variations are modest on InfoSeek, with slightly larger effects for $K$ and $\lambda$ and smaller effects for $U_{\mathrm{STAR}}$ and $T_{\max}$. This pattern supports the view that \textit{CLIMB}'s gains are driven by its structural design, complementary pooling and confidence-guided refinement, rather than by delicate hyperparameter tuning.

\subsection{Qualitative Results}
\begin{figure*}[ht]
  \centering
    \includegraphics[width=0.95\linewidth]{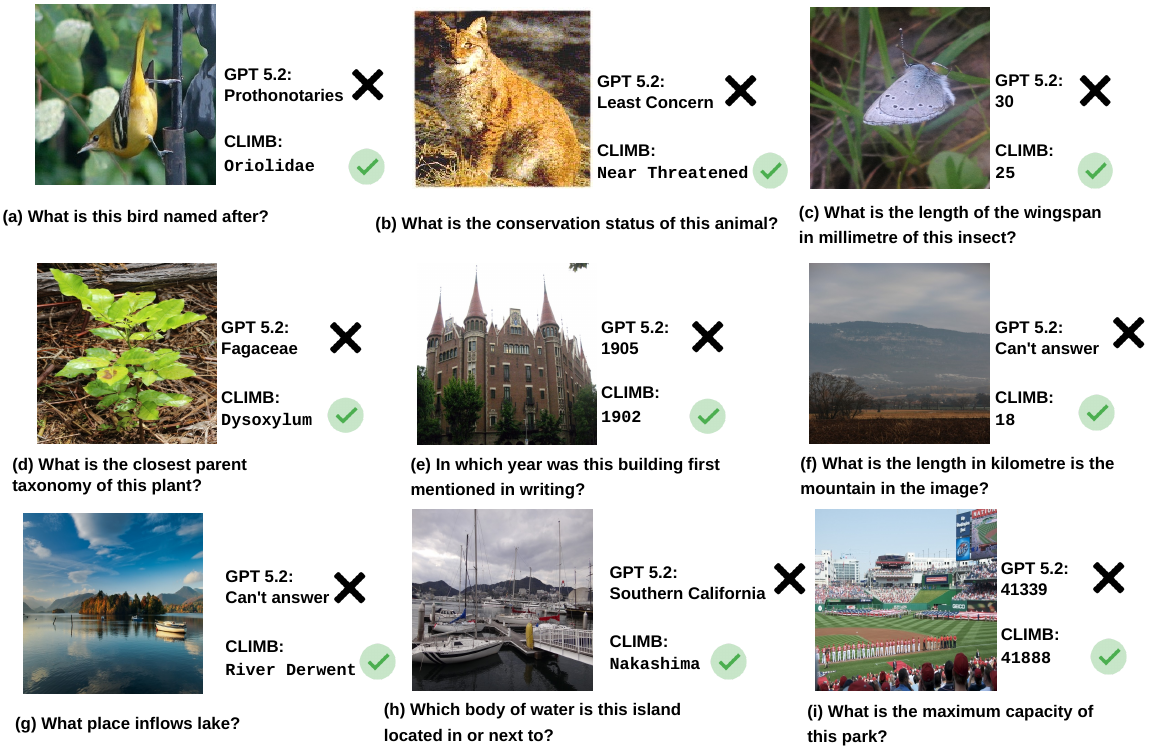}
  \vspace{-3pt}
\caption{Question–Image pairs from the InfoSeek, shown with answers produced by GPT-5.2 and \textit{CLIMB}.}

    \vspace{-10pt}
    \label{fig:qualitative}
\end{figure*}
In addition to the quantitative results, we also include qualitative comparison using GPT-5.2 as a reference model in Figure \ref{fig:qualitative}. Showing side-by-side outputs from GPT-5.2 and \textit{CLIMB} helps illustrate how complementary evidence and confidence-guided refinement affect the final answers beyond what is visible from aggregate metrics alone.

\section{Conclusion}
We introduced \textit{CLIMB}, a training-free framework for multimodal retrieval-augmented generation that strengthens knowledge-intensive visual question answering by building a complementary evidence pool and refining answers under a confidence acceptance rule. Rather than scaling model size or retrieval depth, \textit{CLIMB} focuses on which passages to trust and when to update. On both datasets, it steadily surpasses the baselines, with ablations and sensitivity studies indicating that the gains come from this structural design. We hope this view encourages further work on verifiability-aware, complementarity-driven multimodal RAG.

\section*{Limitations}
This work focuses on knowledge-intensive visual question answering with external Wikipedia-style textual evidence. While this setting is widely used for evaluating multimodal RAG, it does not exhaust all multimodal retrieval scenarios. Extending CLIMB to video, multi-image reasoning, or domain-specific corpora may require adapting the retrieval interface and evidence scoring prompts. The confidence score in CLIMB is used as a relative signal for accepting or rejecting answer updates. It should therefore be interpreted as an evidence-grounded refinement criterion rather than as a calibrated probability of correctness. Future work could investigate more specialized confidence estimators while keeping the framework training-free.

\section*{Acknowledgments}
This work was supported in part by NSF grants 2212302, 2212301, 2212303, AFOSR 23RT0630, and NIH 2R01HL127661.

\bibliography{custom}

@inproceedings{caffagni2024revolution,
    title = "The Revolution of Multimodal Large Language Models: A Survey",
    author = "Caffagni, Davide  and
      Cocchi, Federico  and
      Barsellotti, Luca  and
      Moratelli, Nicholas  and
      Sarto, Sara  and
      Baraldi, Lorenzo  and
      Baraldi, Lorenzo  and
      Cornia, Marcella  and
      Cucchiara, Rita",
    editor = "Ku, Lun-Wei  and
      Martins, Andre  and
      Srikumar, Vivek",
    booktitle = "Findings of the Association for Computational Linguistics: ACL 2024",
    month = aug,
    year = "2024",
    address = "Bangkok, Thailand",
    publisher = "Association for Computational Linguistics",
    url = "https://aclanthology.org/2024.findings-acl.807/",
    doi = "10.18653/v1/2024.findings-acl.807",
    pages = "13590--13618"
}

@inproceedings{wu2023multimodal,
  title={Multimodal large language models: A survey},
  author={Wu, Jiayang and Gan, Wensheng and Chen, Zefeng and Wan, Shicheng and Yu, Philip S},
  booktitle={2023 IEEE International Conference on Big Data (BigData)},
  pages={2247--2256},
  year={2023},
  organization={IEEE}
}

@article{yin2024survey,
   title={A survey on multimodal large language models},
   volume={11},
   ISSN={2053-714X},
   url={http://dx.doi.org/10.1093/nsr/nwae403},
   DOI={10.1093/nsr/nwae403},
   number={12},
   journal={National Science Review},
   publisher={Oxford University Press (OUP)},
   author={Yin, Shukang and Fu, Chaoyou and Zhao, Sirui and Li, Ke and Sun, Xing and Xu, Tong and Chen, Enhong},
   year={2024},
   month=Nov }

@inproceedings{ye2024mplug,
  title={MPLUG-OwI2: Revolutionizing multi-modal large language model with modality collaboration},
  author={Ye, Qinghao and Xu, Haiyang and Ye, Jiabo and Yan, Ming and Hu, Anwen and Liu, Haowei and Qian, Qi and Zhang, Ji and Huang, Fei},
  booktitle={2024 IEEE/CVF Conference on Computer Vision and Pattern Recognition (CVPR)},
  pages={13040--13051},
  year={2024},
  organization={IEEE}
}

@inproceedings{li2023blip,
  title={Blip-2: Bootstrapping language-image pre-training with frozen image encoders and large language models},
  author={Li, Junnan and Li, Dongxu and Savarese, Silvio and Hoi, Steven},
  booktitle={International conference on machine learning},
  pages={19730--19742},
  year={2023},
  organization={PmLR}
}

@inproceedings{chen2023can,
    title = "Can Pre-trained Vision and Language Models Answer Visual Information-Seeking Questions?",
    author = "Chen, Yang  and
      Hu, Hexiang  and
      Luan, Yi  and
      Sun, Haitian  and
      Changpinyo, Soravit  and
      Ritter, Alan  and
      Chang, Ming-Wei",
    editor = "Bouamor, Houda  and
      Pino, Juan  and
      Bali, Kalika",
    booktitle = "Proceedings of the 2023 Conference on Empirical Methods in Natural Language Processing",
    month = dec,
    year = "2023",
    address = "Singapore",
    publisher = "Association for Computational Linguistics",
    url = "https://aclanthology.org/2023.emnlp-main.925/",
    doi = "10.18653/v1/2023.emnlp-main.925",
    pages = "14948--14968"
}

@inproceedings{mensink2023encyclopedic,
  title={Encyclopedic vqa: Visual questions about detailed properties of fine-grained categories},
  author={Mensink, Thomas and Uijlings, Jasper and Castrejon, Lluis and Goel, Arushi and Cadar, Felipe and Zhou, Howard and Sha, Fei and Araujo, Andr{\'e} and Ferrari, Vittorio},
  booktitle={2023 IEEE/CVF International Conference on Computer Vision (ICCV)},
  pages={3090--3101},
  year={2023},
  organization={IEEE}
}

@article{lewis2020retrieval,
  title={Retrieval-augmented generation for knowledge-intensive nlp tasks},
  author={Patrick Lewis and Ethan Perez and Aleksandra Piktus and Fabio Petroni and Vladimir Karpukhin and Naman Goyal and Heinrich Küttler and Mike Lewis and Wen-tau Yih and Tim Rocktäschel and Sebastian Riedel and Douwe Kiela},
  journal={Advances in neural information processing systems},
  volume={33},
  pages={9459--9474},
  year={2020}
}

@inproceedings{caffagni2024wiki,
  title={Wiki-llava: Hierarchical retrieval-augmented generation for multimodal llms},
  author={Caffagni, Davide and Cocchi, Federico and Moratelli, Nicholas and Sarto, Sara and Cornia, Marcella and Baraldi, Lorenzo and Cucchiara, Rita},
  booktitle={2024 IEEE/CVF Conference on Computer Vision and Pattern Recognition Workshops (CVPRW)},
  pages={1818--1826},
  year={2024},
  organization={IEEE}
}

@inproceedings{yan2024echosight,
    title = "{E}cho{S}ight: Advancing Visual-Language Models with {W}iki Knowledge",
    author = "Yan, Yibin  and
      Xie, Weidi",
    editor = "Al-Onaizan, Yaser  and
      Bansal, Mohit  and
      Chen, Yun-Nung",
    booktitle = "Findings of the Association for Computational Linguistics: EMNLP 2024",
    month = nov,
    year = "2024",
    address = "Miami, Florida, USA",
    publisher = "Association for Computational Linguistics",
    url = "https://aclanthology.org/2024.findings-emnlp.83/",
    doi = "10.18653/v1/2024.findings-emnlp.83",
    pages = "1538--1551"
}

@inproceedings{cocchi2025augmenting,
  title={Augmenting multimodal llms with self-reflective tokens for knowledge-based visual question answering},
  author={Cocchi, Federico and Moratelli, Nicholas and Cornia, Marcella and Baraldi, Lorenzo and Cucchiara, Rita},
  booktitle={2025 IEEE/CVF Conference on Computer Vision and Pattern Recognition (CVPR)},
  pages={9199--9209},
  year={2025},
  organization={IEEE}
}

@inproceedings{zhao2023retrieving,
    title = "Retrieving Multimodal Information for Augmented Generation: A Survey",
    author = "Zhao, Ruochen  and
      Chen, Hailin  and
      Wang, Weishi  and
      Jiao, Fangkai  and
      Do, Xuan Long  and
      Qin, Chengwei  and
      Ding, Bosheng  and
      Guo, Xiaobao  and
      Li, Minzhi  and
      Li, Xingxuan  and
      Joty, Shafiq",
    editor = "Bouamor, Houda  and
      Pino, Juan  and
      Bali, Kalika",
    booktitle = "Findings of the Association for Computational Linguistics: EMNLP 2023",
    month = dec,
    year = "2023",
    address = "Singapore",
    publisher = "Association for Computational Linguistics",
    url = "https://aclanthology.org/2023.findings-emnlp.314/",
    doi = "10.18653/v1/2023.findings-emnlp.314",
    pages = "4736--4756"
}

@inproceedings{liu2025hm,
  title={Hm-rag: Hierarchical multi-agent multimodal retrieval augmented generation},
  author={Liu, Pei and Liu, Xin and Yao, Ruoyu and Liu, Junming and Meng, Siyuan and Wang, Ding and Ma, Jun},
  booktitle={Proceedings of the 33rd ACM international conference on multimedia},
  pages={2781--2790},
  year={2025}
}

@inproceedings{arefeen2024irag,
  title={irag: Advancing rag for videos with an incremental approach},
  author={Arefeen, Md Adnan and Debnath, Biplob and Uddin, Md Yusuf Sarwar and Chakradhar, Srimat},
  booktitle={Proceedings of the 33rd ACM International Conference on Information and Knowledge Management},
  pages={4341--4348},
  year={2024}
}

@inproceedings{mckinzie2024mm1,
  title={Mm1: methods, analysis and insights from multimodal llm pre-training},
  author={Brandon McKinzie and Zhe Gan and Jean-Philippe Fauconnier and Sam Dodge and Bowen Zhang and Philipp Dufter and Dhruti Shah and Xianzhi Du and Futang Peng and Floris Weers and Anton Belyi and Haotian Zhang and Karanjeet Singh and Doug Kang and Ankur Jain and Hongyu Hè and Max Schwarzer and Tom Gunter and Xiang Kong and Aonan Zhang and Jianyu Wang and Chong Wang and Nan Du and Tao Lei and Sam Wiseman and Guoli Yin and Mark Lee and Zirui Wang and Ruoming Pang and Peter Grasch and Alexander Toshev and Yinfei Yang},
  booktitle={European Conference on Computer Vision},
  pages={304--323},
  year={2024},
  organization={Springer}
}

@inproceedings{sun2024emu,
  title={Emu: Generative pretraining in multimodality},
  author={Sun, Quan and Yu, Qiying and Cui, Yufeng and Zhang, Fan and Zhang, Xiaosong and Wang, Yueze and Gao, Hongcheng and Liu, Jingjing and Huang, Tiejun and Wang, Xinlong},
  booktitle={International conference on learning representations},
  volume={2024},
  pages={12352--12380},
  year={2024}
}

@article{liu2023visual,
  title={Visual instruction tuning},
  author={Liu, Haotian and Li, Chunyuan and Wu, Qingyang and Lee, Yong Jae},
  journal={Advances in neural information processing systems},
  volume={36},
  pages={34892--34916},
  year={2023}
}

@inproceedings{radford2021learning,
  title={Learning transferable visual models from natural language supervision},
  author={Alec Radford and Jong Wook Kim and Chris Hallacy and Aditya Ramesh and Gabriel Goh and Sandhini Agarwal and Girish Sastry and Amanda Askell and Pamela Mishkin and Jack Clark and Gretchen Krueger and Ilya Sutskever},
  booktitle={International conference on machine learning},
  pages={8748--8763},
  year={2021},
  organization={PmLR}
}

@article{alayrac2022flamingo,
  title={Flamingo: a visual language model for few-shot learning},
  author={Jean-Baptiste Alayrac and Jeff Donahue and Pauline Luc and Antoine Miech and Iain Barr and Yana Hasson and Karel Lenc and Arthur Mensch and Katie Millican and Malcolm Reynolds and Roman Ring and Eliza Rutherford and Serkan Cabi and Tengda Han and Zhitao Gong and Sina Samangooei and Marianne Monteiro and Jacob Menick and Sebastian Borgeaud and Andrew Brock and Aida Nematzadeh and Sahand Sharifzadeh and Mikolaj Binkowski and Ricardo Barreira and Oriol Vinyals and Andrew Zisserman and Karen Simonyan},
  journal={Advances in neural information processing systems},
  volume={35},
  pages={23716--23736},
  year={2022}
}

@inproceedings{changpinyo2021conceptual,
  title={Conceptual 12m: Pushing web-scale image-text pre-training to recognize long-tail visual concepts},
  author={Changpinyo, Soravit and Sharma, Piyush and Ding, Nan and Soricut, Radu},
  booktitle={2021 IEEE/CVF Conference on Computer Vision and Pattern Recognition (CVPR)},
  pages={3557--3567},
  year={2021},
  organization={IEEE}
}

@article{gadre2023datacomp,
  title={Datacomp: In search of the next generation of multimodal datasets},
  author={Samir Yitzhak Gadre and Gabriel Ilharco and Alex Fang and Jonathan Hayase and Georgios Smyrnis and Thao Nguyen and Ryan Marten and Mitchell Wortsman and Dhruba Ghosh and Jieyu Zhang and Eyal Orgad and Rahim Entezari and Giannis Daras and Sarah Pratt and Vivek Ramanujan and Yonatan Bitton and Kalyani Marathe and Stephen Mussmann and Richard Vencu and Mehdi Cherti and Ranjay Krishna and Pang Wei Koh and Olga Saukh and Alexander Ratner and Shuran Song and Hannaneh Hajishirzi and Ali Farhadi and Romain Beaumont and Sewoong Oh and Alex Dimakis and Jenia Jitsev and Yair Carmon and Vaishaal Shankar and Ludwig Schmidt},
  journal={Advances in Neural Information Processing Systems},
  volume={36},
  pages={27092--27112},
  year={2023}
}

@inproceedings{deitke2024molmo,
  title={Molmo and pixmo: Open weights and open data for state-of-the-art vision-language models},
  author={Matt Deitke and Christopher Clark and Sangho Lee and Rohun Tripathi and Yue Yang and Jae Sung Park and Mohammadreza Salehi and Niklas Muennighoff and Kyle Lo and Luca Soldaini and Jiasen Lu and Taira Anderson and Erin Bransom and Kiana Ehsani and Huong Ngo and YenSung Chen and Ajay Patel and Mark Yatskar and Chris Callison-Burch and Andrew Head and Rose Hendrix and Favyen Bastani and Eli VanderBilt and Nathan Lambert and Yvonne Chou and Arnavi Chheda and Jenna Sparks and Sam Skjonsberg and Michael Schmitz and Aaron Sarnat and Byron Bischoff and Pete Walsh and Chris Newell and Piper Wolters and Tanmay Gupta and Kuo-Hao Zeng and Jon Borchardt and Dirk Groeneveld and Crystal Nam and Sophie Lebrecht and Caitlin Wittlif and Carissa Schoenick and Oscar Michel and Ranjay Krishna and Luca Weihs and Noah A. Smith and Hannaneh Hajishirzi and Ross Girshick and Ali Farhadi and Aniruddha Kembhavi},
  booktitle={2025 IEEE/CVF Conference on Computer Vision and Pattern Recognition (CVPR)},
  pages={91--104},
  year={2025},
  organization={IEEE}
}

@InProceedings{borgeaud2022improving,
  title = 	 {Improving Language Models by Retrieving from Trillions of Tokens},
  author =       {Borgeaud, Sebastian and Mensch, Arthur and Hoffmann, Jordan and Cai, Trevor and Rutherford, Eliza and Millican, Katie and Van Den Driessche, George Bm and Lespiau, Jean-Baptiste and Damoc, Bogdan and Clark, Aidan and De Las Casas, Diego and Guy, Aurelia and Menick, Jacob and Ring, Roman and Hennigan, Tom and Huang, Saffron and Maggiore, Loren and Jones, Chris and Cassirer, Albin and Brock, Andy and Paganini, Michela and Irving, Geoffrey and Vinyals, Oriol and Osindero, Simon and Simonyan, Karen and Rae, Jack and Elsen, Erich and Sifre, Laurent},
  booktitle = 	 {Proceedings of the 39th International Conference on Machine Learning},
  pages = 	 {2206--2240},
  year = 	 {2022},
  editor = 	 {Chaudhuri, Kamalika and Jegelka, Stefanie and Song, Le and Szepesvari, Csaba and Niu, Gang and Sabato, Sivan},
  volume = 	 {162},
  series = 	 {Proceedings of Machine Learning Research},
  month = 	 {17--23 Jul},
  publisher =    {PMLR},
  url = 	 {https://proceedings.mlr.press/v162/borgeaud22a.html}
}

@inproceedings{guu2020retrieval,
  title={Retrieval augmented language model pre-training},
  author={Guu, Kelvin and Lee, Kenton and Tung, Zora and Pasupat, Panupong and Chang, Mingwei},
  booktitle={International conference on machine learning},
  pages={3929--3938},
  year={2020},
  organization={PMLR}
}

@article{izacard2023atlas,
  title={Atlas: Few-shot learning with retrieval augmented language models},
  author={Izacard, Gautier and Lewis, Patrick and Lomeli, Maria and Hosseini, Lucas and Petroni, Fabio and Schick, Timo and Dwivedi-Yu, Jane and Joulin, Armand and Riedel, Sebastian and Grave, Edouard},
  journal={Journal of Machine Learning Research},
  volume={24},
  number={251},
  pages={1--43},
  year={2023}
}

@inproceedings{asai2024self,
  title={Self-rag: Learning to retrieve, generate, and critique through self-reflection},
  author={Asai, Akari and Wu, Zeqiu and Wang, Yizhong and Sil, Avirup and Hajishirzi, Hannaneh},
  booktitle={International conference on learning representations},
  volume={2024},
  pages={9112--9141},
  year={2024}
}

@article{arslan2024survey,
  title={A Survey on RAG with LLMs},
  author={Arslan, Muhammad and Ghanem, Hussam and Munawar, Saba and Cruz, Christophe},
  journal={Procedia computer science},
  volume={246},
  pages={3781--3790},
  year={2024},
  publisher={Elsevier}
}

@inproceedings{fan2024survey,
  title={A survey on rag meeting llms: Towards retrieval-augmented large language models},
  author={Fan, Wenqi and Ding, Yujuan and Ning, Liangbo and Wang, Shijie and Li, Hengyun and Yin, Dawei and Chua, Tat-Seng and Li, Qing},
  booktitle={Proceedings of the 30th ACM SIGKDD conference on knowledge discovery and data mining},
  pages={6491--6501},
  year={2024}
}

@article{gao2023retrieval,
  title={Retrieval-augmented generation for large language models: A survey},
  author={Gao, Yunfan and Xiong, Yun and Gao, Xinyu and Jia, Kangxiang and Pan, Jinliu and Bi, Yuxi and Dai, Yi and Sun, Jiawei and Wang, Meng and Wang, Haofen},
  journal={arXiv preprint arXiv:2312.10997},
  year={2023}
}

@inproceedings{yang2025knowing,
  title={Knowing You Don't Know: Learning When to Continue Search in Multi-round RAG through Self-Practicing},
  author={Yang, Diji and Zeng, Linda and Rao, Jinmeng and Zhang, Yi},
  booktitle={Proceedings of the 48th International ACM SIGIR Conference on Research and Development in Information Retrieval},
  pages={1305--1315},
  year={2025}
}

@inproceedings{carbonell1998use,
  title={The use of MMR, diversity-based reranking for reordering documents and producing summaries.},
  author={Carbonell, Jaime G and Goldstein, Jade},
  booktitle={SIGIR},
  pages={290941--291025},
  year={1998}
}

@inproceedings{
vicuna,
title={Judging {LLM}-as-a-Judge with {MT}-Bench and Chatbot Arena},
author={Lianmin Zheng and Wei-Lin Chiang and Ying Sheng and Siyuan Zhuang and Zhanghao Wu and Yonghao Zhuang and Zi Lin and Zhuohan Li and Dacheng Li and Eric Xing and Hao Zhang and Joseph E. Gonzalez and Ion Stoica},
booktitle={Thirty-seventh Conference on Neural Information Processing Systems Datasets and Benchmarks Track},
year={2023},
url={https://openreview.net/forum?id=uccHPGDlao}
}

@article{llama3,
  title={The llama 3 herd of models},
  author={Aaron Grattafiori and Abhimanyu Dubey and Abhinav Jauhri and Abhinav Pandey and Abhishek Kadian and Ahmad Al-Dahle and Aiesha Letman and Akhil Mathur and Alan Schelten and Alex Vaughan and Amy Yang and Angela Fan and Anirudh Goyal and Anthony Hartshorn and Aobo Yang and Archi Mitra and Archie Sravankumar and Artem Korenev and Arthur Hinsvark and Arun Rao and Aston Zhang and Aurelien Rodriguez and Austen Gregerson and Ava Spataru and Baptiste Roziere and Bethany Biron and Binh Tang and Bobbie Chern and Charlotte Caucheteux and Chaya Nayak and Chloe Bi and Chris Marra and Chris McConnell and Christian Keller and Christophe Touret and Chunyang Wu and Corinne Wong and Cristian Canton Ferrer and Cyrus Nikolaidis and Damien Allonsius and Daniel Song and Danielle Pintz and Danny Livshits and Danny Wyatt and David Esiobu and Dhruv Choudhary and Dhruv Mahajan and Diego Garcia-Olano and Diego Perino and Dieuwke Hupkes and Egor Lakomkin and Ehab AlBadawy and Elina Lobanova and Emily Dinan and Eric Michael Smith and Filip Radenovic and Francisco Guzmán and Frank Zhang and Gabriel Synnaeve and Gabrielle Lee and Georgia Lewis Anderson and Govind Thattai and Graeme Nail and Gregoire Mialon and Guan Pang and Guillem Cucurell and Hailey Nguyen and Hannah Korevaar and Hu Xu and Hugo Touvron and Iliyan Zarov and Imanol Arrieta Ibarra and Isabel Kloumann and Ishan Misra and Ivan Evtimov and Jack Zhang and Jade Copet and Jaewon Lee and Jan Geffert and Jana Vranes and Jason Park and Jay Mahadeokar and Jeet Shah and Jelmer van der Linde and Jennifer Billock and Jenny Hong and Jenya Lee and Jeremy Fu and Jianfeng Chi and Jianyu Huang and Jiawen Liu and Jie Wang and Jiecao Yu and Joanna Bitton and Joe Spisak and Jongsoo Park and Joseph Rocca and Joshua Johnstun and Joshua Saxe and Junteng Jia and Kalyan Vasuden Alwala and Karthik Prasad and Kartikeya Upasani and Kate Plawiak and Ke Li and Kenneth Heafield and Kevin Stone and Khalid El-Arini and Krithika Iyer and Kshitiz Malik and Kuenley Chiu and Kunal Bhalla and Kushal Lakhotia and Lauren Rantala-Yeary and Laurens van der Maaten and Lawrence Chen and Liang Tan and Liz Jenkins and Louis Martin and Lovish Madaan and Lubo Malo and Lukas Blecher and Lukas Landzaat and Luke de Oliveira and Madeline Muzzi and Mahesh Pasupuleti and Mannat Singh and Manohar Paluri and Marcin Kardas and Maria Tsimpoukelli and Mathew Oldham and Mathieu Rita and Maya Pavlova and Melanie Kambadur and Mike Lewis and Min Si and Mitesh Kumar Singh and Mona Hassan and Naman Goyal and Narjes Torabi and Nikolay Bashlykov and Nikolay Bogoychev and Niladri Chatterji and Ning Zhang and Olivier Duchenne and Onur Çelebi and Patrick Alrassy and Pengchuan Zhang and Pengwei Li and Petar Vasic and Peter Weng and Prajjwal Bhargava and Pratik Dubal and Praveen Krishnan and Punit Singh Koura and Puxin Xu and Qing He and Qingxiao Dong and Ragavan Srinivasan and Raj Ganapathy and Ramon Calderer and Ricardo Silveira Cabral and Robert Stojnic and Roberta Raileanu and Rohan Maheswari and Rohit Girdhar and Rohit Patel and Romain Sauvestre and Ronnie Polidoro and Roshan Sumbaly and Ross Taylor and Ruan Silva and Rui Hou and Rui Wang and Saghar Hosseini and Sahana Chennabasappa and Sanjay Singh and Sean Bell and Seohyun Sonia Kim and Sergey Edunov and Shaoliang Nie and Sharan Narang and Sharath Raparthy and Sheng Shen and Shengye Wan and Shruti Bhosale and Shun Zhang and Simon Vandenhende and Soumya Batra and Spencer Whitman and Sten Sootla and Stephane Collot and Suchin Gururangan and Sydney Borodinsky and Tamar Herman and Tara Fowler and Tarek Sheasha and Thomas Georgiou and Thomas Scialom and Tobias Speckbacher and Todor Mihaylov and Tong Xiao and Ujjwal Karn and Vedanuj Goswami and Vibhor Gupta and Vignesh Ramanathan and Viktor Kerkez and Vincent Gonguet and Virginie Do and Vish Vogeti and Vítor Albiero and Vladan Petrovic and Weiwei Chu and Wenhan Xiong and Wenyin Fu and Whitney Meers and Xavier Martinet and Xiaodong Wang and Xiaofang Wang and Xiaoqing Ellen Tan and Xide Xia and Xinfeng Xie and Xuchao Jia and Xuewei Wang and Yaelle Goldschlag and Yashesh Gaur and Yasmine Babaei and Yi Wen and Yiwen Song and Yuchen Zhang and Yue Li and Yuning Mao and Zacharie Delpierre Coudert and Zheng Yan and Zhengxing Chen and Zoe Papakipos and Aaditya Singh and Aayushi Srivastava and Abha Jain and Adam Kelsey and Adam Shajnfeld and Adithya Gangidi and Adolfo Victoria and Ahuva Goldstand and Ajay Menon and Ajay Sharma and Alex Boesenberg and Alexei Baevski and Allie Feinstein and Amanda Kallet and Amit Sangani and Amos Teo and Anam Yunus and Andrei Lupu and Andres Alvarado and Andrew Caples and Andrew Gu and Andrew Ho and Andrew Poulton and Andrew Ryan and Ankit Ramchandani and Annie Dong and Annie Franco and Anuj Goyal and Aparajita Saraf and Arkabandhu Chowdhury and Ashley Gabriel and Ashwin Bharambe and Assaf Eisenman and Azadeh Yazdan and Beau James and Ben Maurer and Benjamin Leonhardi and Bernie Huang and Beth Loyd and Beto De Paola and Bhargavi Paranjape and Bing Liu and Bo Wu and Boyu Ni and Braden Hancock and Bram Wasti and Brandon Spence and Brani Stojkovic and Brian Gamido and Britt Montalvo and Carl Parker and Carly Burton and Catalina Mejia and Ce Liu and Changhan Wang and Changkyu Kim and Chao Zhou and Chester Hu and Ching-Hsiang Chu and Chris Cai and Chris Tindal and Christoph Feichtenhofer and Cynthia Gao and Damon Civin and Dana Beaty and Daniel Kreymer and Daniel Li and David Adkins and David Xu and Davide Testuggine and Delia David and Devi Parikh and Diana Liskovich and Didem Foss and Dingkang Wang and Duc Le and Dustin Holland and Edward Dowling and Eissa Jamil and Elaine Montgomery and Eleonora Presani and Emily Hahn and Emily Wood and Eric-Tuan Le and Erik Brinkman and Esteban Arcaute and Evan Dunbar and Evan Smothers and Fei Sun and Felix Kreuk and Feng Tian and Filippos Kokkinos and Firat Ozgenel and Francesco Caggioni and Frank Kanayet and Frank Seide and Gabriela Medina Florez and Gabriella Schwarz and Gada Badeer and Georgia Swee and Gil Halpern and Grant Herman and Grigory Sizov and Guangyi and Zhang and Guna Lakshminarayanan and Hakan Inan and Hamid Shojanazeri and Han Zou and Hannah Wang and Hanwen Zha and Haroun Habeeb and Harrison Rudolph and Helen Suk and Henry Aspegren and Hunter Goldman and Hongyuan Zhan and Ibrahim Damlaj and Igor Molybog and Igor Tufanov and Ilias Leontiadis and Irina-Elena Veliche and Itai Gat and Jake Weissman and James Geboski and James Kohli and Janice Lam and Japhet Asher and Jean-Baptiste Gaya and Jeff Marcus and Jeff Tang and Jennifer Chan and Jenny Zhen and Jeremy Reizenstein and Jeremy Teboul and Jessica Zhong and Jian Jin and Jingyi Yang and Joe Cummings and Jon Carvill and Jon Shepard and Jonathan McPhie and Jonathan Torres and Josh Ginsburg and Junjie Wang and Kai Wu and Kam Hou U and Karan Saxena and Kartikay Khandelwal and Katayoun Zand and Kathy Matosich and Kaushik Veeraraghavan and Kelly Michelena and Keqian Li and Kiran Jagadeesh and Kun Huang and Kunal Chawla and Kyle Huang and Lailin Chen and Lakshya Garg and Lavender A and Leandro Silva and Lee Bell and Lei Zhang and Liangpeng Guo and Licheng Yu and Liron Moshkovich and Luca Wehrstedt and Madian Khabsa and Manav Avalani and Manish Bhatt and Martynas Mankus and Matan Hasson and Matthew Lennie and Matthias Reso and Maxim Groshev and Maxim Naumov and Maya Lathi and Meghan Keneally and Miao Liu and Michael L. Seltzer and Michal Valko and Michelle Restrepo and Mihir Patel and Mik Vyatskov and Mikayel Samvelyan and Mike Clark and Mike Macey and Mike Wang and Miquel Jubert Hermoso and Mo Metanat and Mohammad Rastegari and Munish Bansal and Nandhini Santhanam and Natascha Parks and Natasha White and Navyata Bawa and Nayan Singhal and Nick Egebo and Nicolas Usunier and Nikhil Mehta and Nikolay Pavlovich Laptev and Ning Dong and Norman Cheng and Oleg Chernoguz and Olivia Hart and Omkar Salpekar and Ozlem Kalinli and Parkin Kent and Parth Parekh and Paul Saab and Pavan Balaji and Pedro Rittner and Philip Bontrager and Pierre Roux and Piotr Dollar and Polina Zvyagina and Prashant Ratanchandani and Pritish Yuvraj and Qian Liang and Rachad Alao and Rachel Rodriguez and Rafi Ayub and Raghotham Murthy and Raghu Nayani and Rahul Mitra and Rangaprabhu Parthasarathy and Raymond Li and Rebekkah Hogan and Robin Battey and Rocky Wang and Russ Howes and Ruty Rinott and Sachin Mehta and Sachin Siby and Sai Jayesh Bondu and Samyak Datta and Sara Chugh and Sara Hunt and Sargun Dhillon and Sasha Sidorov and Satadru Pan and Saurabh Mahajan and Saurabh Verma and Seiji Yamamoto and Sharadh Ramaswamy and Shaun Lindsay and Shaun Lindsay and Sheng Feng and Shenghao Lin and Shengxin Cindy Zha and Shishir Patil and Shiva Shankar and Shuqiang Zhang and Shuqiang Zhang and Sinong Wang and Sneha Agarwal and Soji Sajuyigbe and Soumith Chintala and Stephanie Max and Stephen Chen and Steve Kehoe and Steve Satterfield and Sudarshan Govindaprasad and Sumit Gupta and Summer Deng and Sungmin Cho and Sunny Virk and Suraj Subramanian and Sy Choudhury and Sydney Goldman and Tal Remez and Tamar Glaser and Tamara Best and Thilo Koehler and Thomas Robinson and Tianhe Li and Tianjun Zhang and Tim Matthews and Timothy Chou and Tzook Shaked and Varun Vontimitta and Victoria Ajayi and Victoria Montanez and Vijai Mohan and Vinay Satish Kumar and Vishal Mangla and Vlad Ionescu and Vlad Poenaru and Vlad Tiberiu Mihailescu and Vladimir Ivanov and Wei Li and Wenchen Wang and Wenwen Jiang and Wes Bouaziz and Will Constable and Xiaocheng Tang and Xiaojian Wu and Xiaolan Wang and Xilun Wu and Xinbo Gao and Yaniv Kleinman and Yanjun Chen and Ye Hu and Ye Jia and Ye Qi and Yenda Li and Yilin Zhang and Ying Zhang and Yossi Adi and Youngjin Nam and Yu and Wang and Yu Zhao and Yuchen Hao and Yundi Qian and Yunlu Li and Yuzi He and Zach Rait and Zachary DeVito and Zef Rosnbrick and Zhaoduo Wen and Zhenyu Yang and Zhiwei Zhao and Zhiyu Ma},
  journal={arXiv preprint arXiv:2407.21783},
  year={2024}
}

@inproceedings{llava1.5,
  title={Improved baselines with visual instruction tuning},
  author={Liu, Haotian and Li, Chunyuan and Li, Yuheng and Lee, Yong Jae},
  booktitle={2024 IEEE/CVF Conference on Computer Vision and Pattern Recognition (CVPR)},
  pages={26286--26296},
  year={2024},
  organization={IEEE}
}

@article{abdin2024phi,
  title={Phi-4 technical report},
  author={Marah Abdin and Jyoti Aneja and Harkirat Behl and Sébastien Bubeck and Ronen Eldan and Suriya Gunasekar and Michael Harrison and Russell J. Hewett and Mojan Javaheripi and Piero Kauffmann and James R. Lee and Yin Tat Lee and Yuanzhi Li and Weishung Liu and Caio C. T. Mendes and Anh Nguyen and Eric Price and Gustavo de Rosa and Olli Saarikivi and Adil Salim and Shital Shah and Xin Wang and Rachel Ward and Yue Wu and Dingli Yu and Cyril Zhang and Yi Zhang},
  journal={arXiv preprint arXiv:2412.08905},
  year={2024}
}

@article{li2024llava,
  title={Llava-onevision: Easy visual task transfer},
  author={Bo Li and Yuanhan Zhang and Dong Guo and Renrui Zhang and Feng Li and Hao Zhang and Kaichen Zhang and Peiyuan Zhang and Yanwei Li and Ziwei Liu and Chunyuan Li},
  journal={arXiv preprint arXiv:2408.03326},
  year={2024}
}

@inproceedings{lerner2024cross,
  title={Cross-modal retrieval for knowledge-based visual question answering},
  author={Lerner, Paul and Ferret, Olivier and Guinaudeau, Camille},
  booktitle={European Conference on Information Retrieval},
  pages={421--438},
  year={2024},
  organization={Springer}
}

@article{qi2024rora,
  title={Rora-vlm: Robust retrieval-augmented vision language models},
  author={Qi, Jingyuan and Xu, Zhiyang and Shao, Rulin and Chen, Yang and Di, Jin and Cheng, Yu and Wang, Qifan and Huang, Lifu},
  journal={arXiv preprint arXiv:2410.08876},
  year={2024}
}

@inproceedings{
gao2026beyond,
title={Beyond Explicit Edges: Robust Reasoning over Noisy and Sparse Knowledge Graphs},
author={Hang Gao and Dimitris N. Metaxas},
booktitle={Forty-third International Conference on Machine Learning},
year={2026},
url={https://openreview.net/forum?id=5F0ddHN3TP}
}

@article{gao2026pooling,
  title={Pooling and Semantic Shift: The Fundamental Challenges in Long Text Embedding and Retrieval},
  author={Gao, Hang and Xu, Wujiang and Mei, Kai and Metaxas, Dimitris N},
  journal={arXiv preprint arXiv:2603.21437},
  year={2026}
}

@article{gao2024vector,
  title={Vector Retrieval with Similarity and Diversity: How Hard Is It?},
  author={Gao, Hang and Deng, Dong and Zhang, Yongfeng},
  journal={arXiv preprint arXiv:2407.04573},
  year={2024}
}
\appendix
\section*{APPENDIX}
\section{Implementation and Metric Details}
\label{sec:supp_implementation}

This section provides additional details on how we instantiate \textit{CLIMB} in practice. The system is decomposed into five deterministic modules: (i) corpus indexing and micro-pool construction, (ii) R/E/C evidence scoring, (iii) answer generation with explicit evidence conditioning, (iv) a verifiability-oriented confidence function, and (v) facet diagnosis with an iterative confidence climb. All components operate purely at inference time; no finetuning or gradient-based updates are used.

\noindent
\textbf{System Overview. }
Given an input image $I$, question $q$, and a text-only corpus $\mathcal{D}$, the pipeline proceeds as follows:
\begin{enumerate}
    \item Encode all corpus passages with CLIP ViT-L/14 and build a FAISS index using inner-product similarity.
    \item For a query image $I$, retrieve an initial candidate set and construct a complementary micro-pool of size $K{=}5$ using a Maximal Marginal Relevance (MMR) objective.
    \item For each passage in the pool, compute three interpretable scores: relevance $R$, evidence specificity $E$, and cross-modal consistency $C$, and aggregate them into a single ranking signal $\phi$.
    \item Select the top-$M$ passages under $\phi$, generate a short textual answer $a$, and compute a scalar confidence $U\in[0,1]$ based on verifiability criteria.
    \item If $U$ is below a target threshold $U_{\star}$, diagnose under-supported facets, refresh scores inside the fixed micro-pool, and repeat the generation--confidence step. An update is accepted only if it strictly increases $U$, yielding a monotone confidence climb with at most $T_{\max}{=}3$ rounds.
\end{enumerate}

\noindent
\textbf{Complementary Micro-pool Construction. }
We first embed all corpus passages using CLIP ViT-L/14 and store the vectors in a FAISS index. At query time, the image $I$ is encoded into a normalized embedding, and we compute cosine similarity against all passage embeddings. Let $\mathrm{rel}(t)$ denote the similarity between the image embedding and passage $t$, and let $\mathrm{sim}(t,u)$ denote the cosine similarity between two passages $t$ and $u$ in the text-embedding space.

To avoid redundancy in the evidence context, we construct a micro-pool $\mathcal{P}$ with an MMR-style objective. Starting from the most relevant passage, we iteratively add one passage at a time by maximizing
\begin{equation}
\mathrm{MMR}(t)
\,=\,
\lambda\,\mathrm{rel}(t)
-(1-\lambda)\,\max_{u\in S}\mathrm{sim}(t,u),
\end{equation}
where $S$ is the set of already selected passages, and $\lambda\in[0,1]$ balances query fit and redundancy. In our implementation we use $K{=}5$ and set $\lambda$ in the range $[0.5,0.6]$. With cached embeddings, the overall complexity is $O(NK)$ for $N$ initial candidates.

\noindent
\textbf{R/E/C Evidence Scoring. }
Once the micro-pool $\mathcal{P}$ is fixed, we assign each passage $t\in\mathcal{P}$ three scores in $[1,5]$:
\begin{itemize}
    \item Relevance $R(t)$: how directly and comprehensively the passage addresses the question $q$.
    \item Evidence Specificity $E(t)$: how rich the passage is in concrete, verifiable details (named entities, dates, numeric values, etc.).
    \item Cross-modal Consistency $C(t)$: how well the passage aligns with visual cues in $I$ (category, attributes, location, or other scene constraints).
\end{itemize}
These scores are produced by a structured evaluator that returns a JSON object with fields \texttt{R}, \texttt{E}, \texttt{C}, and a brief textual rationale. All calls use temperature $0$, which makes the scoring highly stable across runs. We then normalize the scores to $[0,1]$ and compute a unified ranking signal
\begin{equation}
\phi(t)
\,=\, \alpha_R\,R(t) + \alpha_E\,E(t) + \alpha_C\,C(t),
\end{equation}
with fixed weights $\alpha_R,\alpha_E,\alpha_C$ chosen on a small development split. The top-$M$ passages under $\phi$ form the initial working subset $\mathcal{S}_0$.

\noindent
\textbf{Answer Generation and Confidence Estimation. }
For a given working subset $\mathcal{S}$, we generate an answer in two explicit steps:
\begin{enumerate}
    \item Key information extraction: the system first identifies key facts in the concatenated evidence that directly address the question (e.g., entity names, dates, numeric values). The output is a short, structured summary of only the relevant facts.
    \item Short answer decoding: conditioned on the extracted key information (and optionally the image), the system produces the shortest possible answer (typically a single word or short phrase) without explanations or hedging phrases.
\end{enumerate}
We then compute a scalar confidence $U\in[0,1]$ for the tuple $(I,q,\mathcal{S},a)$. The confidence function is implemented as a verifier that returns a JSON object with fields such as:
\begin{itemize}
    \item \texttt{confidence}: overall confidence score in $[0,1]$,
    \item \texttt{completeness}: coverage of the question by the evidence,
    \item \texttt{reliability}: quality and specificity of the supporting evidence,
    \item \texttt{reasoning}: a short explanation of the assessment.
\end{itemize}
If JSON parsing fails, we fall back to a conservative default. In all reported experiments, we use the \texttt{confidence} field as the scalar $U$ that drives the confidence-climb decision.

\noindent
\textbf{Facet Diagnosis and Iterative Confidence Climb}
Many errors arise when the current subset $\mathcal{S}_t$ misses a key facet of the question. Instead of issuing new retrieval calls, we perform facet diagnosis within the fixed pool $\mathcal{P}$. Given $(I,q,\mathcal{S}_t,a_t)$, we extract at most two short facet phrases
\begin{equation}
\mathcal{G}_t = \{g_1,g_2\},
\end{equation}
each constrained to be no longer than a few tokens, describing missing aspects that prevent full support. A relaxed mode is used by default: if the answer already roughly addresses the question, the diagnoser returns “no significant gaps” and the loop terminates.

When non-empty $\mathcal{G}_t$ is returned, we re-score passages inside $\mathcal{P}$ by combining the original score $\phi(t)$ with a facet–text match score $\psi(t\mid\mathcal{G}_t)$, which mixes lexical overlap (BM25-style) and embedding similarity:
\begin{equation}
\phi'(t) = \phi(t) + \gamma\,\psi(t\mid\mathcal{G}_t),
\end{equation}
where $\gamma$ is a small positive constant. The next subset $\mathcal{S}_{t+1}$ is obtained by taking the top-$M$ passages under $\phi'$, and we regenerate an answer $a_{t+1}$ with an updated confidence $U_{t+1}$. We enforce a monotone acceptance rule:
\begin{equation}
\begin{aligned}
U_{t+1} > U_t 
&\;\Rightarrow\; \text{accept and continue}, \\
U_{t+1} \le U_t 
&\;\Rightarrow\; \text{stop}.
\end{aligned}
\end{equation}
In practice, we set $U_{\star}{=}0.85$ as a target confidence and cap the number of refinement rounds by $T_{\max}{=}3$.

\section{Datasets}
This section provides additional details about the datasets used in our experiments. We describe the
construction process, data sources, and annotation characteristics of Encyclopedic-VQA and InfoSeek,
with an emphasis on the types of visual grounding and factual reasoning they require. These details are
included to clarify the benchmark-specific assumptions and to ensure the experimental setup can be
fully reproduced.

\noindent
\textbf{Encyclopedic-VQA.}
Encyclopedic-VQA is a large-scale, entity-centric benchmark designed to test fine-grained factual
reasoning in multimodal settings. Each instance consists of (i) an image depicting a real-world entity,
(ii) a question targeting a specific factual attribute, and (iii) a short answer extracted verbatim from a
Wikipedia article. Images are sourced from several curated recognition datasets, including iNaturalist
2021, Google Landmarks, and other domain-specific corpora, ensuring substantial variation in species,
landmarks, geological formations, artworks, and historical objects.  

Each image is linked to a corresponding Wikipedia entity page, and the dataset constructs its
retrieval corpus by segmenting roughly two million pages into paragraph-level units. These segments
constitute the complete evidence pool for all experiments, and retrieval is strictly restricted to this
controlled corpus.  

The question design emphasizes factual specificity. Many questions require identifying fine-grained
attributes such as biological taxonomy (e.g., “Order?”, “Genus?”), geographical information (“Which
river flows through this site?”), historical metadata (“Who inaugurated this building?”), physical
measurements, or conservation status. Importantly, most questions require combining visual grounding
(e.g., identifying the depicted species) with factual lookup (e.g., locating the required line in the
Wikipedia entry). The official evaluation focuses on the single-hop setting, where answers are supported
by a single paragraph from the article, allowing controlled analysis of evidence retrieval.

\noindent
\textbf{InfoSeek.}
InfoSeek serves as a complementary benchmark emphasizing broader question styles and larger-scale
variability. It contains approximately 1.3M automatically generated image--question--answer triplets,
augmented with a curated set of human-authored questions for quality assurance. Entities span more
than ten thousand Wikipedia pages, including species, buildings, artworks, geographic regions, public
figures, and artifacts.  

Each example is paired with a textual corpus consisting of around one hundred thousand aligned
Wikipedia pages. Compared to Encyclopedic-VQA, the InfoSeek knowledge base is smaller but covers a
wider breadth of topics. The dataset intentionally introduces greater lexical and structural diversity in
question templates—for example, describing the same fact using different surface forms, or mixing
attribute, relational, temporal, and descriptive queries.  

InfoSeek is especially challenging because the model must (i) correctly recognize the visual entity
from the image (fine-grained categories such as subspecies, architectural variants, instrument models),
(ii) identify the relevant page among many visually similar candidates, and (iii) pinpoint the correct fact
within the retrieved document. The evaluation protocol follows the official validation split and restricts
retrieval to the provided corpus to ensure comparability.

\noindent
\textbf{Comparison and Usage.}
Together, the two datasets provide complementary perspectives on multimodal knowledge-intensive
reasoning. Encyclopedic-VQA emphasizes controlled factual grounding with a very large retrieval
corpus and fine-grained entity alignment, while InfoSeek emphasizes large-scale variability, surface
diversity in questions, and robustness to paraphrases and distributional shifts. 

\section{Prompts}
The templates used to instantiate the R/E/C scoring, confidence estimation, and facet-diagnosis interfaces.
\subsection{R/E/C Evidence Judge Prompt}
\begin{quote}
You are given an image, a question, and one retrieved passage. Score the passage for answering the question.

Question: \{question\}

Passage: \{passage\}

Evaluate the retrieved content on the following dimensions (1--5 scale):

1. Relevance (R): How directly and comprehensively does the content address the evaluation question?

\quad - 5: Directly and fully answers the question

\quad - 4: Mostly relevant with minor gaps

\quad - 3: Partially relevant but missing key aspects

\quad - 2: Loosely related, limited usefulness

\quad - 1: Not relevant or tangential

2. Evidence Specificity (E): How precise, detailed, and factual is the information?

\quad - 5: Highly specific facts, dates, names, or concrete details

\quad - 4: Specific information with some details

\quad - 3: Moderately specific, some generalizations

\quad - 2: Vague or general statements

\quad - 1: Very vague or lacks substantive information

3. Cross-modal Consistency (C): Is the passage consistent with the visual content of the image?

\quad - 5: Perfectly consistent and complementary

\quad - 4: Mostly consistent with minor discrepancies

\quad - 3: Generally consistent but some inconsistencies

\quad - 2: Some contradictions or inconsistencies

\quad - 1: Contradictory or clearly inconsistent

Return only a JSON object in the following format:
\begin{verbatim}
{"R": integer from 1 to 5,
 "E": integer from 1 to 5,
 "C": integer from 1 to 5}
\end{verbatim}
\end{quote}

\subsection{Confidence Estimation Prompt}
\begin{quote}
You are given an image, a question, a set of selected evidence passages, and a candidate answer. Estimate how well the answer is supported by the image and evidence.

Question: \{question\}

Selected evidence:
\{selected\_evidence\}

Candidate answer: \{answer\}

Evaluation Criteria:

1. Completeness: Do the evidences contain sufficient information to fully answer the question?

2. Relevance: How directly related are the evidences to the question?

3. Reliability: Are the evidences specific, factual, and trustworthy?

4. Consistency: Do the evidences support each other, or are there contradictions?

5. Answer Quality: Does the answer correctly interpret and synthesize the evidences?

Evaluate the answer using three output scores:

1. Coverage: Are the factual claims in the answer supported by the selected evidence?

2. Consistency: Is the answer consistent with the image and with the selected evidence?

3. Specificity: Is the answer concrete enough to be verified, rather than vague or evasive?

Assign each output score a value between 0 and 1. 
Return only a JSON object in the following format:
\begin{verbatim}
{"coverage": float from 0 to 1,
 "consistency": float from 0 to 1,
 "specificity": float from 0 to 1}
\end{verbatim}
\end{quote}
\subsection{Facet Diagnosis Prompt}

\begin{quote}
You are given an image, a question, the currently selected evidence passages, and the current answer. Identify which aspect of the answer is missing, weakly supported, or needs additional evidence.

Question: \{question\}

Current selected evidence:
\{selected\_evidence\}

Current answer: \{answer\}

List at most two short facet phrases that describe missing or under-supported information needed to better answer the question. A facet should be a concise phrase, such as an entity attribute, date, location, measurement, taxonomy, conservation status, or historical fact.

If the current answer is already sufficiently supported, or if no meaningful missing facet can be identified, return an empty list.

Return only a JSON object in the following format:
\begin{verbatim}
{"facets": ["short facet phrase 1",
            "short facet phrase 2"]}
\end{verbatim}
\end{quote}

\section{Failure Case}
\begin{table*}[t]
\centering
\small
\setlength{\tabcolsep}{4pt}
\renewcommand{\arraystretch}{1.18}
\begin{tabular}{p{0.12\linewidth}p{0.22\linewidth}p{0.20\linewidth}p{0.16\linewidth}p{0.21\linewidth}}
\hline
\textbf{Step} & \textbf{Evidence focus} & \textbf{Diagnosis / action} & \textbf{Model answer} & \textbf{Failure signal} \\
\hline
\multicolumn{5}{p{0.94\linewidth}}{
\textbf{Question:} ``What is the closest parent taxonomy of this animal?'' 
\quad \textbf{Gold answer:} \textit{Loriinae}.
} \\
\hline
Initial generation &
Evidence mentions \textit{Loriini} as a tribe and emphasizes the broader parrot family \textit{Psittaculidae}. &
Generate an answer from the initial selected subset. &
\textit{Psittaculidae} &
The answer is related and specific, but it is a broader ancestor rather than the closest parent. \\
\hline
Facet diagnosis &
Current evidence supports taxonomy but does not clearly resolve the nearest parent. &
Diagnose the missing facet as ``taxonomic parentage''. &
-- &
The diagnosis is reasonable, so the error is not simply a failed facet proposal. \\
\hline
Pool-internal refresh &
Additional passages contain adjacent taxonomy levels: family \textit{Psittaculidae}, subfamily \textit{Loriinae}, and tribe \textit{Loriini}. &
Refresh evidence inside the fixed pool and regenerate. &
\textit{Psittaculidae} &
Evidence is more complete and internally consistent, but still permits the wrong granularity. \\
\hline
Acceptance &
The refined answer appears supported by concrete taxonomy terms. &
Accept the update because estimated confidence increases. &
Final: \textit{Psittaculidae}.&
Confidence can increase despite incorrectness because specificity and consistency do not guarantee closest-parent resolution. \\
\hline
\end{tabular}
\caption{A failure analysis from InfoSeek.}
\label{tab:loriini-failure}
\end{table*}
\end{document}